\documentclass[oribibl]{llncs}
\usepackage{makeidx}      
\usepackage{graphicx}     
\usepackage{mathrsfs}     
\usepackage[ruled,linesnumbered,vlined]{algorithm2e}
\usepackage{color}
\usepackage{amsfonts}
\usepackage{multirow}

\SetKwProg{Fn}{Function}{}{}
\SetArgSty{textrm}

\begin{document}
%
%
\pagestyle{headings}  

\newcommand{\karen}[1]{\textcolor{red}{{Karen: #1}}}
\newcommand{\edit}[1]{\textcolor{red}{{#1}}}

%
%
\title{Traversing Environments Using Possibility Graphs for Humanoid Robots}

\author{ Michael X. Grey
\and Aaron D. Ames 
\and C. Karen Liu}

\institute{Georgia Institute of Technology, Atlanta GA 30332, USA}

\maketitle

\begin{abstract}
Locomotion for legged robots poses considerable challenges when confronted by obstacles and adverse environments. Footstep planners are typically only designed for one mode of locomotion, but traversing unfavorable environments may require several forms of locomotion to be sequenced together, such as walking, crawling, and jumping. Multi-modal motion planners can be used to address some of these problems, but existing implementations tend to be time-consuming and are limited to quasi-static actions. This paper presents a motion planning method to traverse complex environments using multiple categories of actions. We introduce the concept of the ``Possibility Graph'', which uses high-level approximations of constraint manifolds to rapidly explore the ``possibility'' of actions, thereby allowing lower-level single-action motion planners to be utilized more efficiently. We show that the Possibility Graph can quickly find paths through several different challenging environments which require various combinations of actions in order to traverse.
\end{abstract}

\section{Introduction}

Modern motion planning methods have proven effective at navigating geometric constraint manifolds within high dimensional configurations spaces. This capability is critical for robots to exhibit autonomy in complex real-world environments, because geometric constraints are frequently used to determine the \emph{feasibility} of a physical action and hence are often used as ``feasibility constraints'' which must be satisfied or else the action is considered infeasible. Geometric constraints include requirements such as avoiding obstacles and placing end effectors in appropriate locations. Two common types of motion planners are Probabilistic Roadmaps (PRM) \cite{kavraki:prm} and Rapidly-exploring Random Tree (RRT) \cite{kuffner:connect}. Standard PRM is well-suited for exploring a single \emph{expansive} manifold, as defined in \cite{hsu:expansive}. Constrained Bi-directional RRT (CBiRRT) \cite{berenson:tsr} can effectively handle constraint manifolds whose dimensionality is lower than the configuration space in which it exists.

Standard motion planning methods tend to struggle when a solution needs to traverse numerous topologically distinct constraint manifolds. This occurs most often in hybrid dynamic systems where the ``mode'' of the system alters its constraint manifold. For example, standing on the left foot is a different mode from standing on the right foot for a bipedal robot. The constraint manifolds of these two modes are different, and their intersection is narrow, resulting in a low (sometimes zero) probability of randomly moving from one manifold to the other. Hauser et al. introduced the Multi-modal PRM \cite{hauser:ijrr2009} to address this problem. The primary bottleneck of this method is the combinatorial complexity of sampling and selecting modes, since each footstep taken by the robot represents a mode that must be explored. Additionally, existing implementations of the Multi-modal PRM are limited to quasi-static actions, which broadly eliminates its ability to utilize the dynamic capabilities of a robot system.

\begin{figure}
  \centering
  \includegraphics[width=0.98\textwidth]{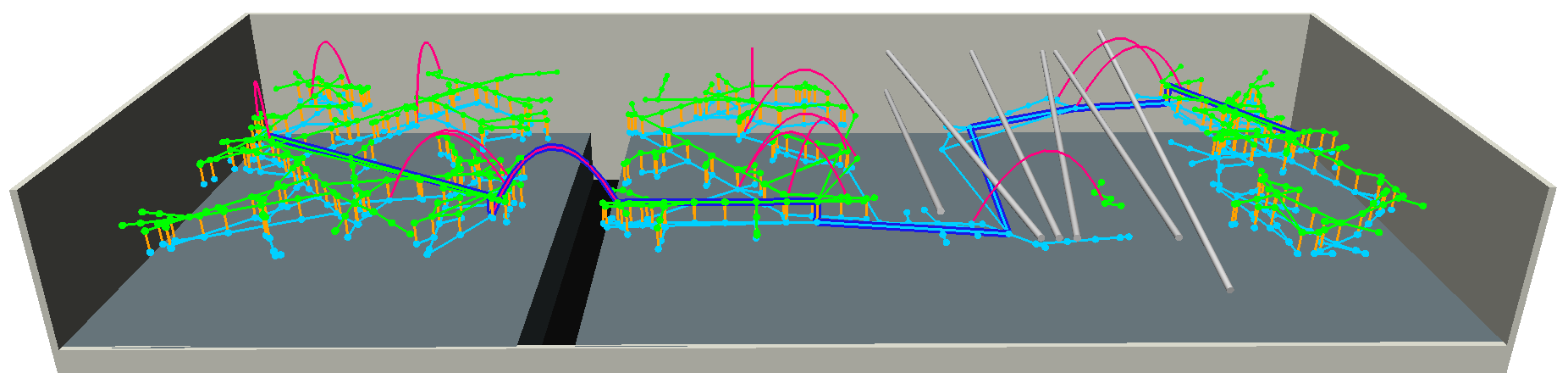}
  
  (a)
  
  \vspace{2mm}
  \includegraphics[width=0.98\textwidth]{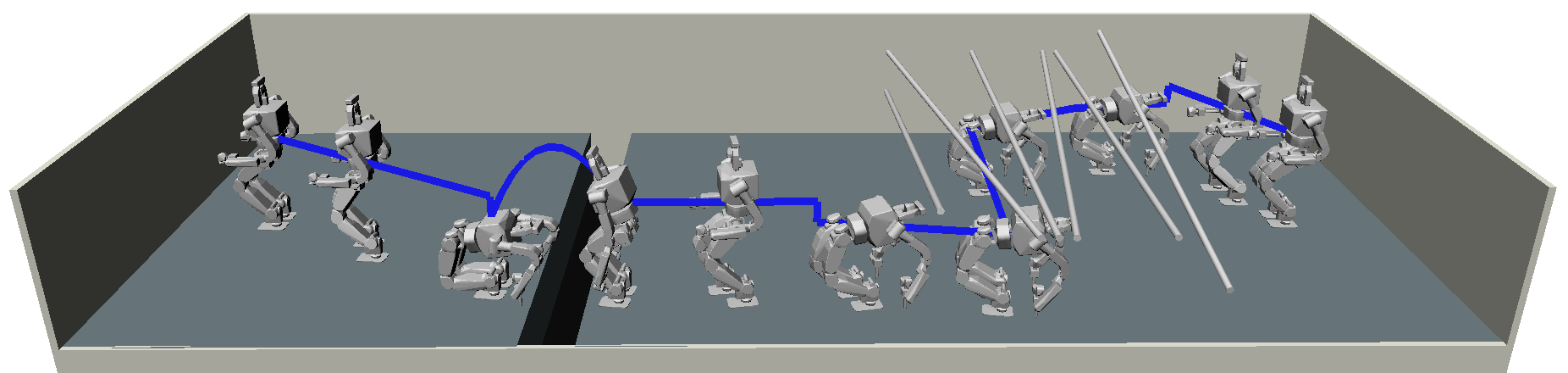}
  (b)
  
  \caption{\label{fig:hallway}The robot is tasked with traversing from the right side of a hallway to the left side. It must navigate underneath bars which are positioned at various angles, and then must jump across a gap. (a) The graph explored the space of the hallway until a solution was found. Green edges are walking actions, light blue edges are crawling actions, and fuchsia arcs are jumping actions. (b) Snapshots show the plan in action.}
\end{figure}

In contrast to motion planning methods, standard footstep planners are able to rapidly generate footsteps and walking trajectories without spending time exploring the constraint manifolds of combinatorial modes the way a Multi-modal PRM does. They typically do this by approximating the problem of walking. In \cite{garimort:footsteps, hornung:anytime} this is done using a 2D representation of obstacles and in \cite{garimort:footsteps, hornung:anytime, candido:hierarchical} only a finite set of footstep parameters or action primitives are available to the planner. The two-stage method presented in \cite{pettre:2stages} uses a bounding cylinder to represent the collision geometry of the lower body. All of these estimations inherently limit the completeness of the methods. Moreover, these methods are all limited to a single category of action: bipedal walking.

The use of optimization methods in the motion planning domain has been growing \cite{zucker:chomp, kuindersma:atlas}, especially for walking motions. Nonlinear constrained optimization can elegantly handle the mixed modes and hybrid dynamics \cite{ayonga:hzd} required for walking and crawling. However, they tend to be tailored for generating single behaviors (e.g. a walking behavior that consists of a single- and double-support phase). This is insufficient for traversing a complex environment where a sequence of different types of behaviors is needed. Optimizations methods also tend to be local, making them inappropriate for tackling problems that require a global search.

The goal of this paper is to introduce a new high-level motion planning
method, named the \emph{Possibility Graph}, that can leverage the speed and
efficiency of standard footstep planners with the completeness of randomized
motion planning methods and the dynamics capabilities of optimization-based methods. 
The Possibility Graph is general enough to handle arbitrary categories of actions 
instead of being limited to only walking or stepping primitives. The role of the
Possibility Graph is to quickly explore what actions might be possible
throughout an environment. Different action types are compactly interlaced with each other within the graph, allowing a solution to utilize any action types in any order.
Once a potential route is discovered, lower-level planning methods are used to 
confirm whether the route is truly
feasible. This allows the lower-level (and computationally intensive)
planners to focus their efforts on queries which are likely to achieve
a solution. These queries can be performed in parallel, ensuring that the overall
planning effort does not get bottlenecked by any single challenging step.


The three categories of actions used in this paper are walking,
crawling, and a standing long jump. Figure \ref{fig:hallway} shows
these three actions being utilized in a hallway example. We test the Possibility Graph on various scenarios. In some scenarios,
multiple action categories may be required to reach the goal. We show
that the Possibility Graph works reliably on the order of seconds. Complete solutions tend to generate at faster than 100x real time. This makes Possibility Graphs suitable for online planning. They could
also be incorporated into higher level task planners \cite{grey:garrett:manip} which require numerous high-speed queries.

%

\section{Possibility Graph}

The governing logical principles behind the Possibility Graph have a theoretical grounding in Possibility Theory \cite{dubois:possibility}, but the concepts are intuitive enough that a knowledge of Possibility Theory is not necessary to proceed. It is enough to understand that the \emph{possibility} of any given action instance can be labelled with ``impossible'', ``possible'', or ``indeterminate'' depending on whether the instance satisfies the necessary or sufficient conditions that are assigned to it. The motivation for using a Possibility Graph is two-fold:

\begin{enumerate}
  \item We can design necessary and/or sufficient conditions that can be checked quickly, making expansion of the graph very efficient.
  \item Even though different actions may have constraint manifolds with different dimensionalities, we can design their necessary and/or sufficient conditions to share a common set of parameters, allowing for a single unified graph which combines all actions.
\end{enumerate}

Motion planning methods ordinarily operate by constructing graphs or trees which fully exist within the feasibility constraint manifold of the action they are performing. Remaining within this manifold is a reasonable requirement to place on the graph, because any vertices or edges which step outside of the manifold are, by definition, invalid---which may mean it is physically impossible, or simply harmful to the robot or its surroundings. Unfortunately, for a humanoid robot to remain on the constraint manifold, expensive calls to whole body inverse kinematics solvers (for more on whole body IK, see \cite{sentis:wholebody, sugihara:wholebody, gienger:wholebody}) must be performed. This results in a critical bottleneck if a broad area needs to be explored before finding a solution. By exploring the \emph{possibility} of actions first, we can broadly avoid expensive whole body inverse kinematics queries and easily slide through transition spaces which would otherwise be narrow.

\subsection{Simplifying the Manifold: Sufficient vs. Necessary Conditions}
\label{sec:simplifying}

To construct the Possibility Graph, we must first design sufficient and/or necessary conditions for the feasibility constraint manifold of each action. The two motivations which were mentioned earlier imply that the conditions we create should satisfy two criteria: (1) They should be quick to test, and (2) they should be low dimensional, using as few parameters as is reasonable.

\begin{figure}
  \centering
  \includegraphics[width=0.48\textwidth]{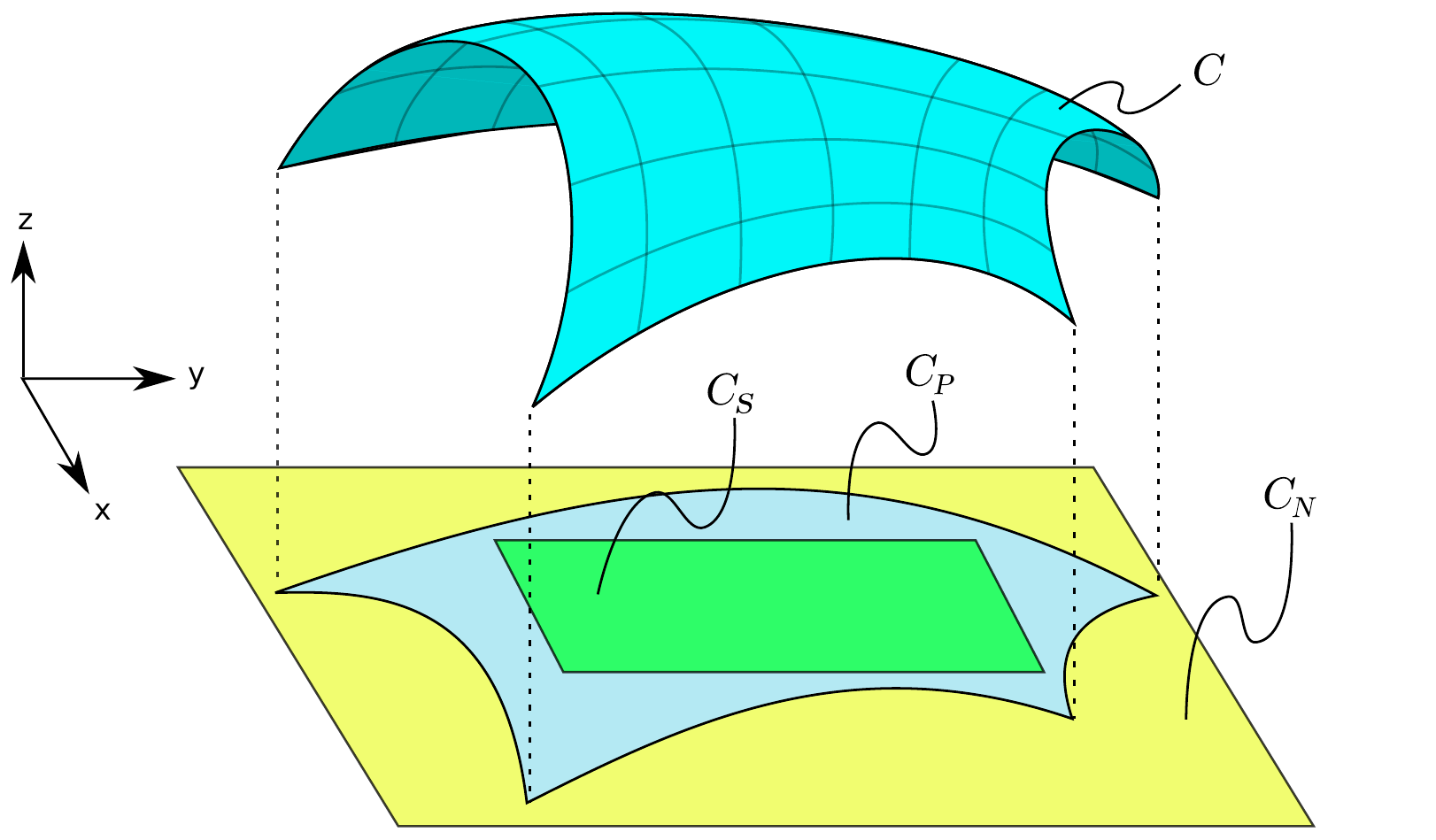}
  
  
  \caption{\label{fig:abstract_manifold}Visual depiction of an abstract constraint manifold, $C$ and its projection. The manifold is projected, $C_P$, from 3D space onto a plane. ``Sufficient'' $C_S$ and ``Necessary'' $C_N$ boundaries are fitted within and around the projection of the manifold.}
\end{figure}

Suppose we have a 2D constraint manifold, $C$, which exists in 3D space, as shown in Fig. \ref{fig:abstract_manifold}(a). Supposing we can directly compute the $z$-value of the manifold given valid $x$ and $y$ values, it makes sense to project this manifold down onto the $xy$-plane. We can call the projection $C_P$. This accomplishes our goal of lower dimensionality.

Even with a flattened-out projection, identifying which points are inside or outside of the manifold may still be costly or difficult, because the boundaries of the projection may be functions that are expensive to compute or hard to fully define. However, suppose a box, circle, or some other simple shape can be fit within the projection such that it is \emph{guaranteed} that every point within the simplified shape also lies within the manifold projection. Such a shape would be a suitable representation of the sufficient condition manifold, $C_S$, for the constraint manifold $C$. Mathematically, this means $C_S \subseteq C_P$. Similarly, if a simple shape, $C_N$, could be fit \emph{around} the projection $C_P$ such that $C_P \subseteq C_N$, then $C_N$ would qualify as the necessary condition manifold. The specific designs of necessary and sufficient manifolds for this paper are discussed in Sec. \ref{sec:action_implementations}.


\begin{figure}
  \centering
  \includegraphics[width=0.97\textwidth]{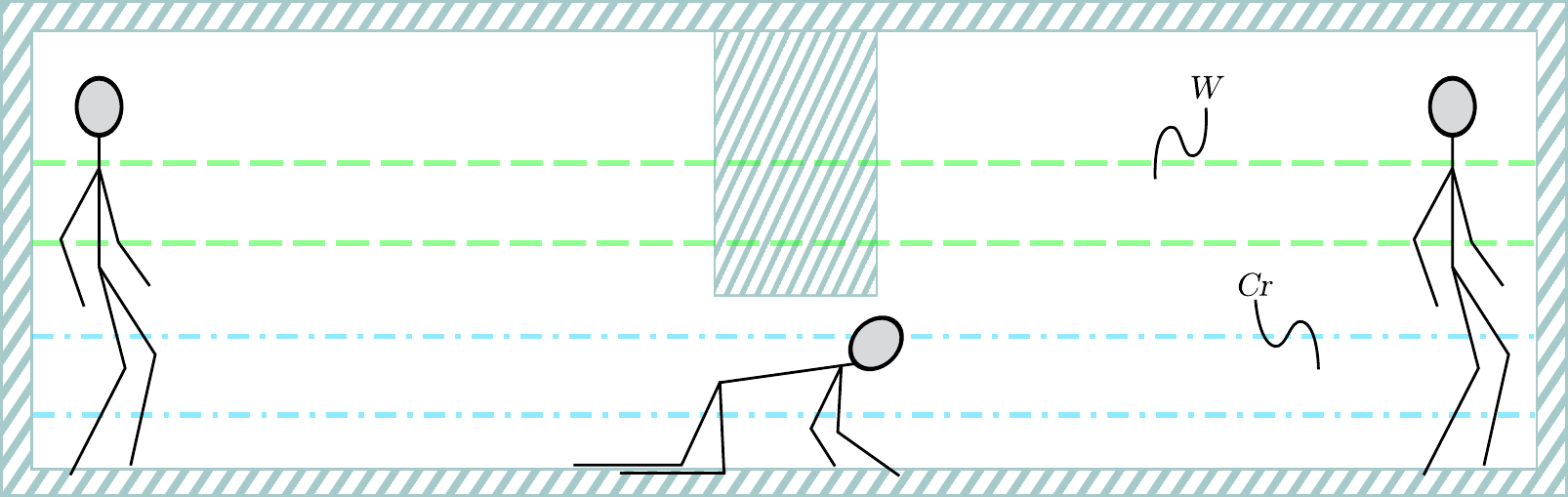}
  
  \vspace{2mm}
  \includegraphics[width=0.97\textwidth]{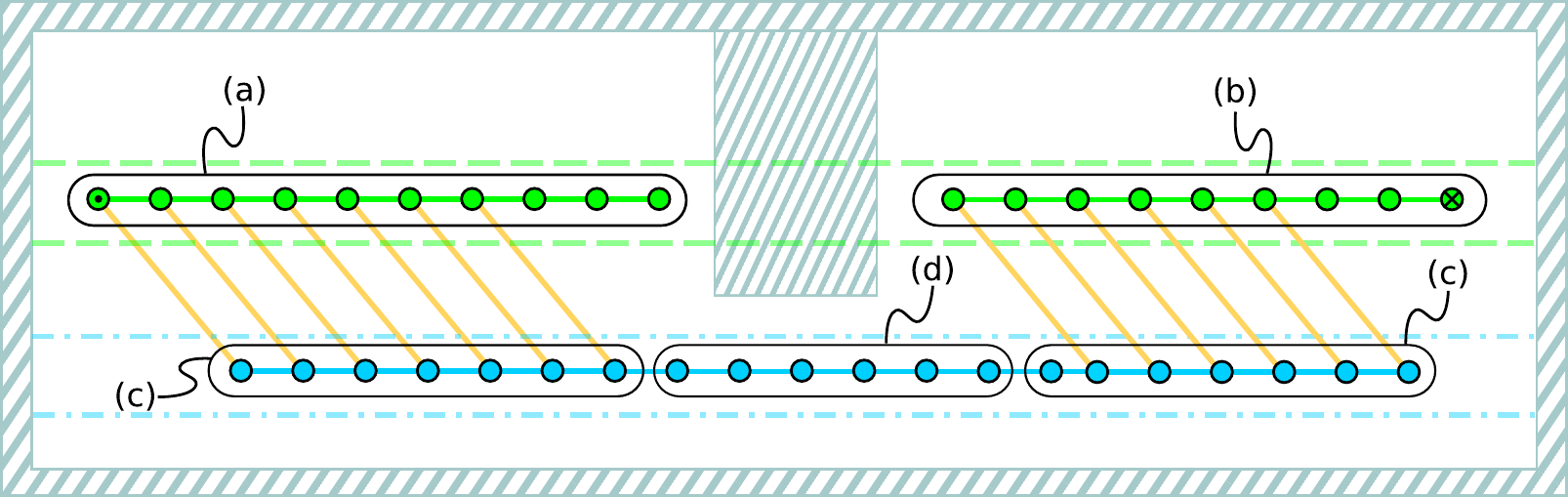}  
  
  \caption{\label{fig:cartoon}Cartoon showing a simple 2D stick-figure example where the stick figure can walk or crawl forward. The graph's vertices represent the $(x,z)$ values of a point fixed to the stick figure's chest. The upper region, marked by $W$ in the top photo, is where walking is valid. The lower region, $Cr$, is where crawling is valid. (a) We extend from the start vertex towards a randomly sampled point in the center. [Alg. \ref{alg:holonomic_growth}, line \ref{alg:holonomic_growth:first_connect}] (b) We extend from the goal vertex towards the last vertex that was created in the previous step. [Alg. \ref{alg:holonomic_growth}, line \ref{alg:holonomic_growth:second_connect}] (c) For each new walking vertex, we create a crawling vertex and connect it to the walking vertex using a transition edge [Alg. \ref{alg:explore}, line \ref{alg:explore:transition}]. For some of the walking vertices, a transition into crawling is not viable due to obstacles. (d) We now extend the crawling subgraphs towards a point that was sampled near the center of the room, and the subgraphs manage to connect [Alg. \ref{alg:explore}, line \ref{alg:explore:growTowards}].}
\end{figure}

\subsection{Explore Possibilities by Expanding the Graph}
\label{sec:explore}

The purpose of the Possibility Graph is to find a feasible action sequence to get from a start point to at least one goal point. We define the contents of the Possibility Graph in Def. \ref{def:posgraph}. The procedure for finding solutions with the graph is described by Alg. \ref{alg:explore}. The graph is initialized with a set of subgraphs, each subgraph consisting solely of either the start point or a goal point. All the graph's vertices are elements of the ``possibility exploration space'', $\mathscr{E}$. The Possibility Graph finds paths through the exploration space by querying each available action to expand the graph in randomized directions within $\mathscr{E}$ (see line \ref{alg:explore:growTowards} of Alg. \ref{alg:explore}). If a query meets at least the necessary conditions of the action, then it will be appended to the graph.

\begin{definition}
\normalfont
\label{def:posgraph}A Possibility Graph is a tuple \\$PG = (\mathrm{\texttt{actions}},\mathscr{E},\Gamma_{PG},Q_{Confirmation})$, where 
\begin{itemize}
\item \texttt{actions} is a set of actions (defined in Table \ref{tab:action}),
\item $\mathscr{E}$ is a space consisting of the union of all the parameters used by the necessary and sufficient conditions of each \texttt{Action},
\item $\Gamma_{PG}$ is a directed graph whose vertices are elements of $\mathscr{E}$,
\item $Q_{Confirmation}$ is a queue which manages confirmation jobs (see Table \ref{tab:action}).
\end{itemize}
\end{definition}

An important feature of our algorithm is the exploration of transitions between various action categories. Each time vertices are added for one action, the other actions will be queried to see if they can transition from it (Alg. \ref{alg:explore} , Line \ref{alg:explore:transition}). This allows different actions to be interlaced with each other within the graph. Each action keeps track of its own exploration by storing a set of subgraphs, $\Gamma_a$, consisting of its own vertices and edges. At the same time, the Possibility Graph maintains the ``master'' graph, $\Gamma_{PG}$, which combines all the subgraphs of all the different action types. The algorithm is illustrated by a toy example in Fig. \ref{fig:cartoon}.

\begin{algorithm}
\caption{\label{alg:explore}Finding a path by exploring possibilities\label{alg:confirmation}}
\Fn{\texttt{FindPath}(start, goals, actions)}{
  $\Gamma_{PG}.V \gets \{\mathrm{start, goals}\}$\;
  Initialize each action graph with the start and goal vertices\;
  $Q_\mathrm{Confirmation}$.\texttt{launchThreads}()\;
  $t \gets 0$\;
  \While{$t < t_{\mathrm{max}}$}{
    \For{$a$ in actions}{
      $\{V_\mathrm{new}, E_\mathrm{new}\} \gets a.\mathrm{\texttt{PerformTransitions}}()$\label{alg:explore:transition}\;
      $p_\mathrm{target} \gets$ \texttt{RandomSample}()\;
      $\{V_\mathrm{new}, E_\mathrm{new}\}$.\texttt{append}($a$.\texttt{GrowTowards}($p_\mathrm{target}$))\label{alg:explore:growTowards}\;
      \For{all $a_\mathrm{other}$ \textbf{not} $a$ in actions}{
        $a_\mathrm{other}.Q_\mathrm{Transition}.\mathrm{\texttt{insert}}(V_\mathrm{new})$\;
      }
      $\{\Gamma_{PG}.V, \Gamma_{PG}.E\}$.\texttt{append}($\{V_\mathrm{new}, E_\mathrm{new}\}$)\;
    }
    
    \For{$g$ in goals}{
      \If{\texttt{Connected}(start, g)}{
        $\Gamma_\mathrm{path} \gets$ \texttt{ShortestPath}(start, $g$)\;
        \If{\texttt{ConfirmPath}($\Gamma_{PG}$, $\Gamma_\mathrm{path}$, $Q_\mathrm{Confirmation}$, actions)}{
          \Return $\Gamma_\mathrm{path}$\;
        }
      }
    }
    $t \gets$ \texttt{CurrentTime}()\;
  }
  \Return null\;
}\vspace{1mm}
\Fn{\texttt{ConfirmPath}($\Gamma_{PG}$, $\Gamma_\mathrm{path}$, $Q_\mathrm{Confirmation}$, actions)}{
  pathConfirmed $\gets$ \textbf{true}\;
  \For{edge in $\Gamma_\mathrm{path}.E$}{
    edgeConfirmed $\gets$ \textbf{false}\;
    \For{$a$ in actions}{
      \If{$a.C_\mathrm{S}(\mathrm{edge})$}{
        edgeConfirmed $\gets$ \textbf{true}\;
      }
      \ElseIf{$a.C_\mathrm{N}(\mathrm{edge})$}{
        $\Gamma_{PG}$.\texttt{remove}(edge)\;
        $Q_\mathrm{Confirmation}$.\texttt{insert}($a$.\texttt{SpawnConfirmationJob}(edge))\;
      }
    }
    \If{\textbf{not} edgeConfirmed}{
      pathConfirmed $\gets$ \textbf{false}\;
    }
  }
  \Return pathConfirmed\;
}
\end{algorithm}

Over time, the Possibility Graph will consist of vertices and edges from various actions interlaced with each other. Some elements of the graph will satisfy the sufficient conditions of their respective actions, but some will only satisfy the necessary conditions. Once the graph contains a path from the start vertex to a goal vertex, we need to inspect the vertices and edges of that path to confirm whether all the path elements are truly feasible. This process is shown in the \texttt{ConfirmPath} function of Alg. \ref{alg:confirmation}. Actions are responsible for spawning ``confirmation jobs'' which are low-level planning routines whose job is to verify whether or not an edge in the possibility graph is truly feasible. These routines are loaded into the Confirmation Queue, $Q_{Confirmation}$. The Confirmation Queue will rotate between running each job to ensure that easy ones are finished promptly while difficult ones do not halt the overall confirmation progress. These jobs are executed on threads which run parallel to the graph expansion and each other. This allows the planner to search for alternative potential solutions when certain edges are difficult to confirm.


\subsection{Extending Action Subgraphs}
\label{sec:actions}

\begin{table}
\caption{\label{tab:action}Definition of an Action}
\begin{tabular}{lll}
\multicolumn{2}{l}{\textbf{All action types}}\vspace{1mm}\\
\texttt{ExtendTowards}($v_0$, $v_1$):   & Create a vertex by moving towards $v_1$ from $v_0$\\
                                        & via this action.\vspace{1mm}\\
$C_N(x)$:                               & Return \textbf{true} if $x$ meets the action's necessary conditions,\\
                                        & otherwise return \textbf{false}. $x$ may be a vertex or an edge.\vspace{1mm}\\
$C_S(x)$:                               & Return \textbf{true} if $x$ meets the action's sufficient conditions,\\
                                        & otherwise return \textbf{false}. $x$ may be a vertex or an edge.\vspace{1mm}\\
\texttt{TransitionFrom}($v$):           & Attempt to return a path that goes from $v$ into the\\
                                        & necessary condition manifold of this action.\vspace{1mm}\\
\texttt{SpawnConfirmationJob}($e$):     & Return a routine (called a confirmation job) which can\\
                                        & examine edge $e$ to ascertain whether it is truly feasible.\vspace{3mm}\\
\multicolumn{2}{l}{\textbf{Holonomic action types}}\vspace{1mm}\\
\texttt{Project}($v$):                  & Attempt to return a point on the necessary\\
                                        & condition manifold which is close to $v$.\vspace{3mm}\\
\multicolumn{2}{l}{\textbf{Nonholonomic action types}}\vspace{1mm}\\
\texttt{ReverseExtend}($v_0$, $v_1$):   & Create a vertex which can arrive at $v_0$ from the\\
                                        & direction of $v_1$ via this action.\\
\texttt{FindLaunchPoint}($v$, $v_1$):   & Return a point, $v_0$, close to $v$ which can be used\\
                                        & in a call to \texttt{ExtendTowards}($v_0$, $v_1$)\vspace{1mm}\\
\texttt{FindLandingPoint}($v$, $v_1$):  & Return a point, $v_0$, close to $v$ which can be used\\
                                        & in a call to \texttt{ReverseExtend}($v_0$, $v_1$)\\
\end{tabular}
\end{table}

\begin{algorithm}
\caption{\label{alg:transitions}Utilizing the Transition Queue}
\Fn{\texttt{Action::PerformTransitions}()}{
  $\{V_\mathrm{new}, E_\mathrm{new}\}  \gets$ \{new VertexQueue, new EdgeQueue\}\;
  $i \gets 0$\;
  \While{$i <$ MaxTransitionsPerCycle}{
    $v \gets \mathrm{\texttt{PopRandom}}(Q_\mathrm{Transitions})$\;
    $\{V_\mathrm{new}, E_\mathrm{new}\}$.\texttt{append}(\texttt{TransitionFrom}(v))\;
    $i \gets i+1$\;
  }
  $\Gamma_a$.\texttt{append}($\{V_\mathrm{new}, E_\mathrm{new}\}$)\;
  \Return $\{V_\mathrm{new}, E_\mathrm{new}\}$\;
}
\end{algorithm}

For the Possibility Graph to explore actions, we need to fully define each action type. Table \ref{tab:action} lays out the implementation-dependent functions which must be engineered for each action type. The functions in that table enable the \texttt{GrowTowards} and \texttt{PerformTransitions} functions to work. \texttt{PerformTransitions} is described in Alg. \ref{alg:transitions}. It simply pulls vertices from other actions out of a queue and attempts to create transitions from those actions to itself. \texttt{GrowTowards} serves two primary roles: (1) expand the graph in new directions, and (2) connect together disjoint subgraphs. The nature of how an action grows will depend on what kind of action it is. For this paper, we have two methods of expanding an action, one for holonomic actions and the other for nonholonomic.

\textbf{Holonomic actions} are expanded using Alg. \ref{alg:holonomic_growth}. When we describe an action as ``holonomic'' in this context, we mean that its sufficient/necessary condition manifold is holonomic. Even if the full feasibility constraint manifold of the action is nonholonomic, it can be treated as holonomic by the Possibility Graph if its necessary/sufficient condition manifold is simplified to be holonomic within the exploration space, $\mathscr{E}$. Alg. \ref{alg:holonomic_growth} shows how the possibilities of holonomic actions are expanded. Importantly, holonomic actions always try to connect disjoint subgraphs together. This procedure is very similar to the growth of a CBiRRT \cite{berenson:tsr}, except that it accommodates numerous directional subgraphs. To avoid having subgraphs needlessly cross over each other, we only extend two at a time: The subgraph who has the vertex closest to the random target is extended towards the target up to some point $v_0$ [Alg. \ref{alg:holonomic_growth}, line \ref{alg:holonomic_growth:first_connect}] (at which point it cannot extend any further); then the second closest subgraph attempts to connect to $v_0$ [Alg. \ref{alg:holonomic_growth}, line \ref{alg:holonomic_growth:second_connect}]. However, if the first subgraph was goal-connected, then the second subgraph must not be (i.e. we skip over the next closest subgraph until we reach one which is not goal-connected), because connecting together two goal-connected subgraphs cannot help in finding a solution.

\textbf{Nonholonomic actions} are expanded in a more complex way than holonomic actions, as shown in Alg. \ref{alg:nonholonomic_growth}. Nonholonomic actions generally cannot move directly towards a goal, so they need to ``line themselves up'' first. We do this by identifying a launch point which is reachable from an existing point on the graph [Alg. \ref{alg:nonholonomic_growth}, line \ref{alg:launch}]. The launch point should be chosen such that it allows the action to land as close to the randomly generated target as possible, so long as the launch point is still reachable from the existing graph. Since nonholonomic actions are also generally direction-dependent, we do the reverse for goal-connected subgraphs [Alg. \ref{alg:nonholonomic_growth}, line \ref{alg:land}]: Pick a landing point which can connect to an existing goal-connected vertex such that it has a viable launch point coming from the direction of the target. Section \ref{sec:standing_long_jump} describes this for the jump action.

\begin{algorithm}
\caption{\label{alg:holonomic_growth}Growing the graph for a holonomic action}
\Fn{\texttt{HolonomicAction::GrowTowards}($p_\mathrm{target}$)}{
  $Q_\mathrm{closest} \gets$ new SortedVertexQueue\;
  \For{$g$ in $\Gamma_a$.SubGraphs}{
    $v \gets$ $g$.\texttt{FindClosestVertexTo}($p_\mathrm{target}$)\;
    $Q_\mathrm{closest}$.\texttt{insert}(dist($v$, $p_\mathrm{target}$), $v$)\;
  }
  $v_0 \gets Q_\mathrm{closest}$.\texttt{pop\_front}()\;
  $\{V_\mathrm{new}, E_\mathrm{new}\} \gets$ \texttt{Connect}($v_0$, $p_\mathrm{target}$)\label{alg:holonomic_growth:first_connect}\;
  $p_\mathrm{target} \gets V_\mathrm{new}$.\texttt{back}()\;
  \If{\texttt{UpstreamFromGoal}($v_0$)}{
    \While{\texttt{UpstreamFromGoal}($Q_\mathrm{closest}$.\texttt{front}()}{
      $Q_\mathrm{closest}$.\texttt{pop\_front}()\;
    }
  }
  $v_1 \gets Q_\mathrm{closest}$.\texttt{pop\_front}()\;
  $\{V_\mathrm{new}, E_\mathrm{new}\}$.\texttt{append}(\texttt{Connect}($v_1, p_\mathrm{target}$))\label{alg:holonomic_growth:second_connect}\;
  $\Gamma_a$.\texttt{append}($\{V_\mathrm{new}, E_\mathrm{new}\}$)\;
  \Return $\{V_\mathrm{new}, E_\mathrm{new}\}$\;
}
\vspace{1mm}
\Fn{\texttt{HolonomicAction::Connect}($v_\mathrm{start}$, $p_\mathrm{target}$)}{
  $\{V_\mathrm{new}, E_\mathrm{new}\}  \gets$ \{new VertexQueue, new EdgeQueue\}\;
  $v_\mathrm{last} \gets v_\mathrm{start}$\;
  $v \gets$ \texttt{ExtendTowards}($v_\mathrm{start}, p_\mathrm{target}$)\;
  $v_p \gets$ \texttt{Project}(v)\;
  \While{$C_\mathrm{N}(v_p)$ \textbf{and} $v \neq p_\mathrm{target}$}{
    edge $\gets$ Edge($v_\mathrm{last}, v_p$)\;
    \If{\textbf{not} $C_\mathrm{N}$(edge)}{
      \textbf{break}\;
    }
    $\{V_\mathrm{new}, E_\mathrm{new}\}$.\texttt{append}($\{v_p, \mathrm{edge}\}$)\;
    $v_\mathrm{last} \gets v_p$\;
    $v \gets$ \texttt{ExtendTowards}($v, p_\mathrm{target}$)\;
    $v_p \gets$ \texttt{Project}($v$)\;
  }
  \Return $\{V_\mathrm{new}, E_\mathrm{new}\}$\;
}
\end{algorithm}

\begin{algorithm}
\caption{\label{alg:nonholonomic_growth}Growing the graph for a nonholonomic action}
\Fn{\texttt{NonholonomicAction::GrowTowards}($p_\mathrm{target}$)}{
  $\{V_\mathrm{new}, E_\mathrm{new}\}  \gets$ \{new VertexQueue, new EdgeQueue\}\;
  $Q_\mathrm{closest} \gets$ new SortedVertexQueue\;
  \For{$v$ in $\Gamma_a.V$}{
    $Q_\mathrm{closest}$.\texttt{insert}(dist($v$, $p_\mathrm{target}$), $v$)\;
  }
  Useful $\gets$ new BooleanArray($\Gamma_a.V$.size(), true)\;
  \For{$v$ in $Q_\mathrm{closest}$}{
    \lIf{\textbf{not} Useful[$v$]}{ continue }
    \If{\textbf{not} \texttt{UpstreamFromGoal}($v$)}{
      $v_\mathrm{launch} \gets$ \texttt{FindLaunchPoint}($v$, $p_\mathrm{target}$)\label{alg:launch}\;
      $v_\mathrm{landing} \gets$ \texttt{ExtendTowards}($v_\mathrm{launch}, p_\mathrm{target}$)\;
      edge $\gets$ Edge($v_\mathrm{launch}$, $v_\mathrm{landing}$)\;
      \If{$C_\mathrm{N}$(edge)}{
        $\{V_\mathrm{new}, E_\mathrm{new}\}.$\texttt{append}($\{v_\mathrm{launch}, v_\mathrm{landing}, \mathrm{edge}\}$)\;
        \texttt{RecursivelySetUpstreamVerticesToFalse}($v$, Useful)\;
      }
    }
    \If{\textbf{not} \texttt{DownstreamFromStart}($v$)}{
      $v_\mathrm{landing} \gets$ \texttt{FindLandingPoint}($v$, $p_\mathrm{target}$)\label{alg:land}\;
      $v_\mathrm{launch} \gets$ \texttt{ReverseExtend}($v_\mathrm{landing}$, $p_\mathrm{target}$)\;
      edge $\gets$ Edge($v_\mathrm{launch}$, $v_\mathrm{landing}$)\;
      \If{$C_\mathrm{N}$(edge)}{
        $\{V_\mathrm{new}, E_\mathrm{new}\}$.\texttt{append}$\{v_\mathrm{launch}, v_\mathrm{landing}, \mathrm{edge}\}$\;
        \texttt{RecursivelySetDownstreamVerticesToFalse}($v$, Useful)\;
      }
    }
  }
  $\Gamma_a$.\texttt{append}($\{V_\mathrm{new}, E_\mathrm{new}\}$)\;
  \Return $\{V_\mathrm{new}, E_\mathrm{new}\}$\;
}
\end{algorithm}

\section{Action Implementations}
\label{sec:action_implementations}

In this paper, we implement three action types to serve as a proof of concept. Two are holonomic and one is nonholonomic. They include walking, crawling, and a standing long jump. We use a model of the DRC-HUBO1 robot, because its kinematic structure is designed to accommodate crawling. The scenarios in which we apply these actions will be discussed in Sec. \ref{sec:experimental}. For the exploration space of the Possibility Graph, $\mathscr{E}$, we use the SE(3) coordinates of a reference frame attached to the robot's pelvis.

\subsection{Walk and Crawl}

\begin{figure}
  \centering
  \includegraphics[width=0.30\textwidth]{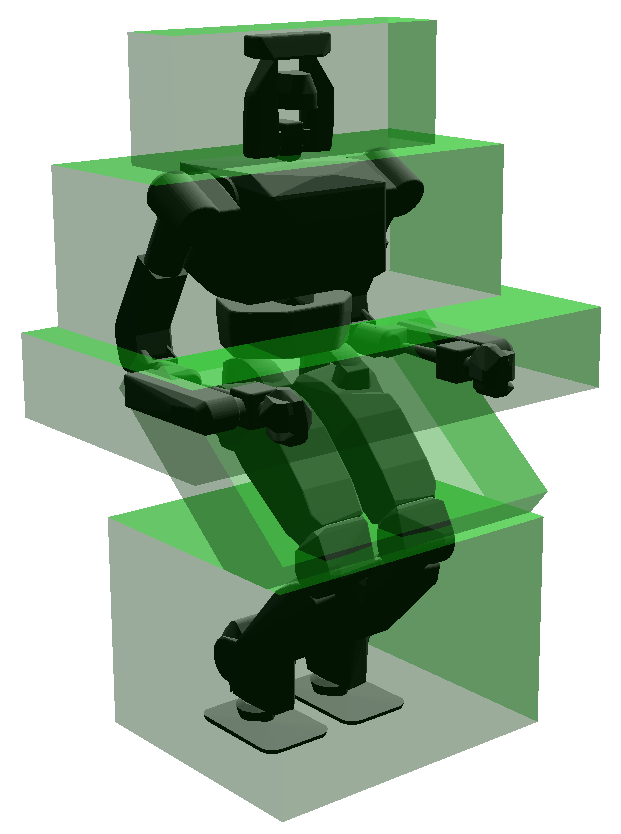}
  \hfil\hfil
  \includegraphics[width=0.40\textwidth]{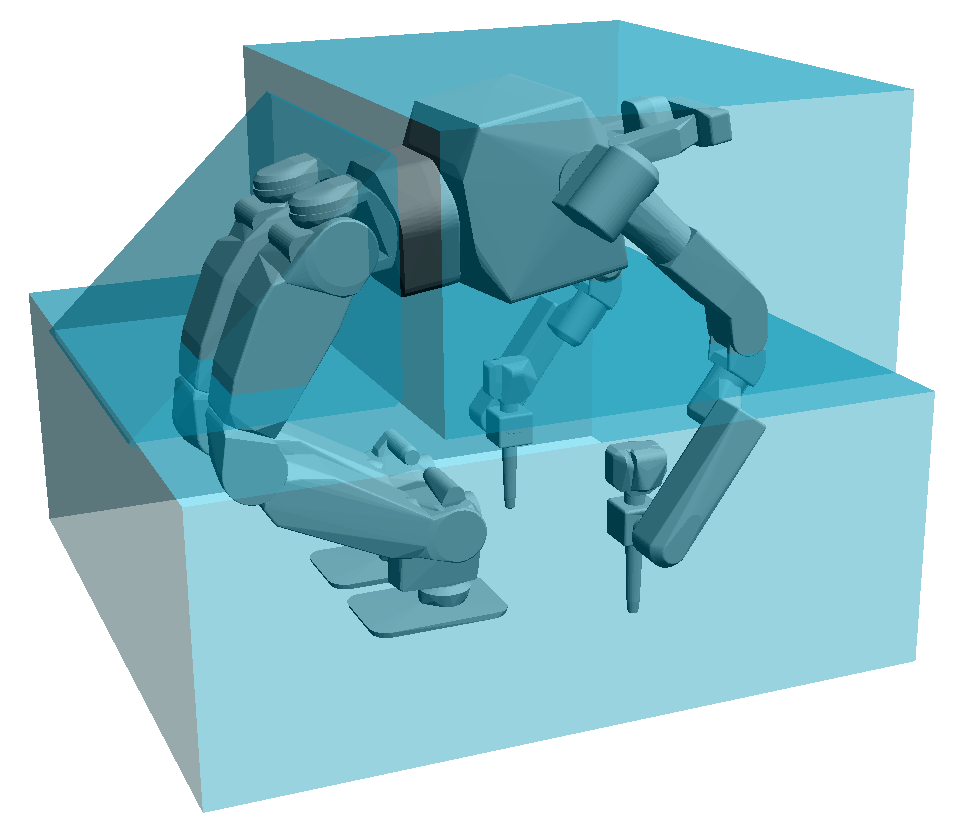}
  
  (a) \hfil \hfil (b)
  
  \caption{\label{fig:sweeps}The nominal configurations used for (a) walking and (b) crawling, with their swept geometries surrounding them.}
\end{figure}

The walking and crawling actions are formulated very similarly to each other. Sufficient conditions for walking and crawling are holonomic, and include these simplifications:

\begin{enumerate}
  \item We use a swept collision geometry, similar to \cite{grey:garrett:manip}. The geometries can be seen in Fig. \ref{fig:sweeps}. These geometries must not be in collision with the environment when given a point in $\mathscr{E}$.
  \item Each point that defines the robot's support polygon must be touching flat ground when the robot is in a ``nominal'' walk/crawl configuration. The nominal configurations can be seen in Fig. \ref{fig:sweeps}.
  \item The root must be in the ``nominal'' orientation of the action (upright for walking and pitched backwards $80^\circ$ for crawling).
\end{enumerate}

The necessary conditions are significantly easier to satisfy:

\begin{enumerate}
  \item We use only the collision geometry of the pelvis, because all other bodies depend on joint parameters which are not included in $\mathscr{E}$.
  \item At least one foot must be able to reach some ground surface.
\end{enumerate}

The \texttt{ExtendTowards}($v_0, v_1$) function simply applies an SE(3) transform which translates and rotates $v_0$ to bring it closer to $v_1$. Changes in rotation should be weighted less than changes in translation in order to have sensible differences between steps. The \texttt{Project} function for these action templates adjusts the height and orientation of the SE(3) input so that it matches the nominal configuration of the action. Translation in $x$/$y$ and rotation along $z$ are unaffected. The \texttt{TransitionFrom} function moves between these actions by generating a simple motion that goes from one nominal configuration to the other. If an edge only meets the necessary conditions of the action, then another planning method (such as Multi-modal PRM) must be generated by the \texttt{SpawnConfirmationJob} function. On the other hand, when the sufficient conditions are satisfied, the final motion for these actions is easily determined by placing footsteps along the specified route through SE(3) and then generating a whole body motion to follow those footsteps. Our sufficient conditions guarantee that it will be possible to generate and follow those footsteps.

\begin{figure}
  \centering

  \includegraphics[width=0.16\textwidth]{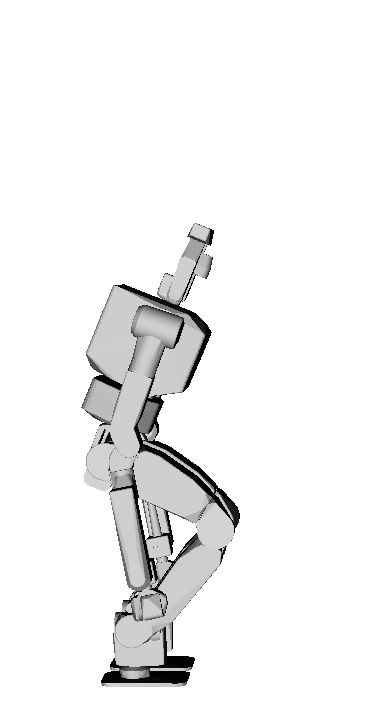}
  \includegraphics[width=0.16\textwidth]{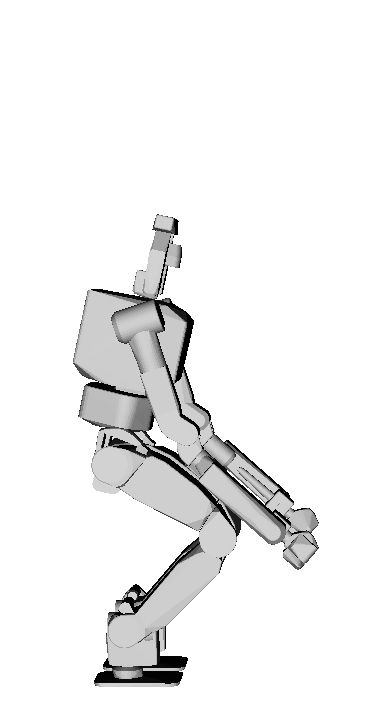}
  \includegraphics[width=0.16\textwidth]{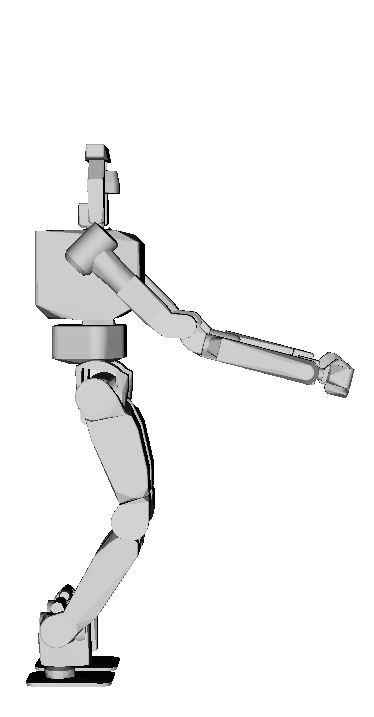}
  \includegraphics[width=0.16\textwidth]{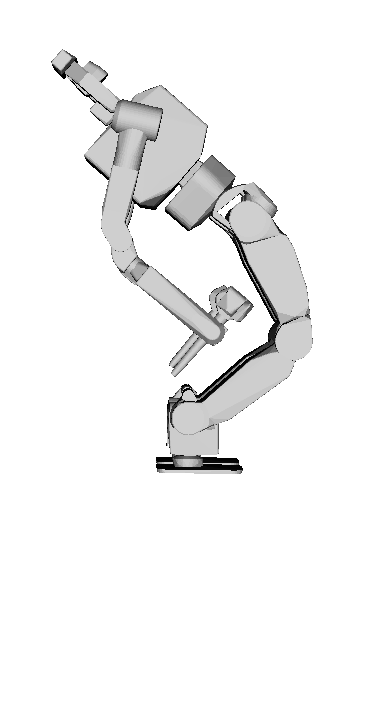}
  \includegraphics[width=0.16\textwidth]{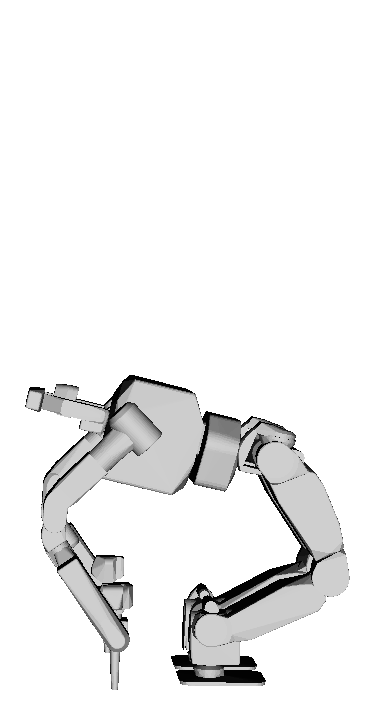}
  \includegraphics[width=0.16\textwidth]{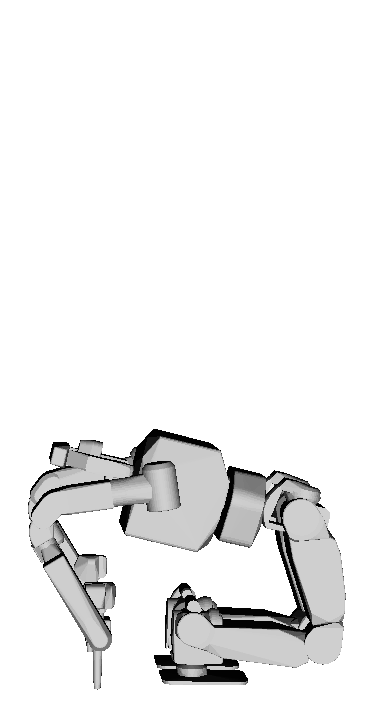}
  
  \caption{\label{fig:jump}An example standing long jump trajectory. The robot begins from a standing configuration, swings its arms, and jumps forward. It plans out its angular momentum so that it is able to land in a crawling configuration. After hitting the ground, it absorbs some of the impact by letting its joints behave elastically.}
\end{figure}

\subsection{Standing Long Jump}
\label{sec:standing_long_jump}
A standing long jump is a forward jump which begins from standing in place and launches forward without taking any steps. Figure \ref{fig:jump} shows an example of a jumping trajectory. We use a standing long jump in this paper for simplicity; future work will include long jumps that take running starts, which can achieve considerably greater range. We provide necessary conditions for the standing long jump but not sufficient conditions. The necessary condition manifold is nonholonomic, and contains the following:

\begin{enumerate}
  \item The vertex that begins the jump must be a valid walk vertex. 
  \item The vertex that finishes the jump must be a valid crawl vertex.
  \item There must be at least one collision-free parabola through $\mathscr{E}$ from the beginning vertex to the finishing vertex. The parabola must follow a feasible jump arc according to the physical limitations of the robot.
\end{enumerate}

The \texttt{TransitionFrom} function for the jump action is trivial, because it always begins from valid walking configurations and ends in valid crawling configurations, therefore the transition function does nothing. The \texttt{ExtendTowards}($v_0, v_1$) function performs a forward jump from $v_0$ to $v_1$. If $v_1$ is too far to reach from $v_0$, then it performs the furthest allowable jump. The \texttt{ReverseExtend}($v_0, v_1$) function instead performs a jump which lands at $v_0$ and begins as close to $v_1$ as the robot's physical limitations allow. The \texttt{FindLaunchingPoint}($v, v_1$) function returns a point, $v_0$, whose translation is the same as $v$ but whose orientation has the robot facing $v_1$; this allows the \texttt{ExtendTowards}($v_0, v_1$) function to bring the robot closer to $v_1$. Conversely, the \texttt{FindLandingPoint}($v, v_1$) function returns a point, $v_0$, whose translation is the same as $v$ but which is facing \emph{away} from $v_1$; this allows the robot to jump towards $v$ from the direction of $v_1$ using \texttt{ReverseExtend}($v_0, v_1$).

The \texttt{SpawnConfirmationJob} function of the jump action is a basic collocation optimization on a boundary value problem. The boundary value constraints are (1) zero initial velocity, (2) a take-off configuration and velocity which will allow the robot to reach its jump target. The objective function of the optimization problem attempts to minimize the accelerations during take-off. While generating the trajectory, we also check that the joint and contact forces required to achieve the trajectory are physically feasible. Trajectories which fail this test are discarded. Once the jump is generated, we can check for collisions along its trajectory. If the jump was successfully generated (i.e. the jumping motion is physically feasible) and is collision-free, then its ``possibility'' status is changed from ``indeterminate'' to ``possible'', and it can be used in a final solution.


\section{Experimental Results}
\label{sec:experimental}

We run performance tests on three scenarios (one of which has three versions). Each performance test is the result of 50 trials. The Possibility Graph is a randomized planner, so the time required for the same trial can vary between runs. We put a 60 second time limit on the planner; if a solution is not found within 60 seconds, we consider it a failed run.

\begin{figure}
  \centering
  \includegraphics[width=0.48\textwidth]{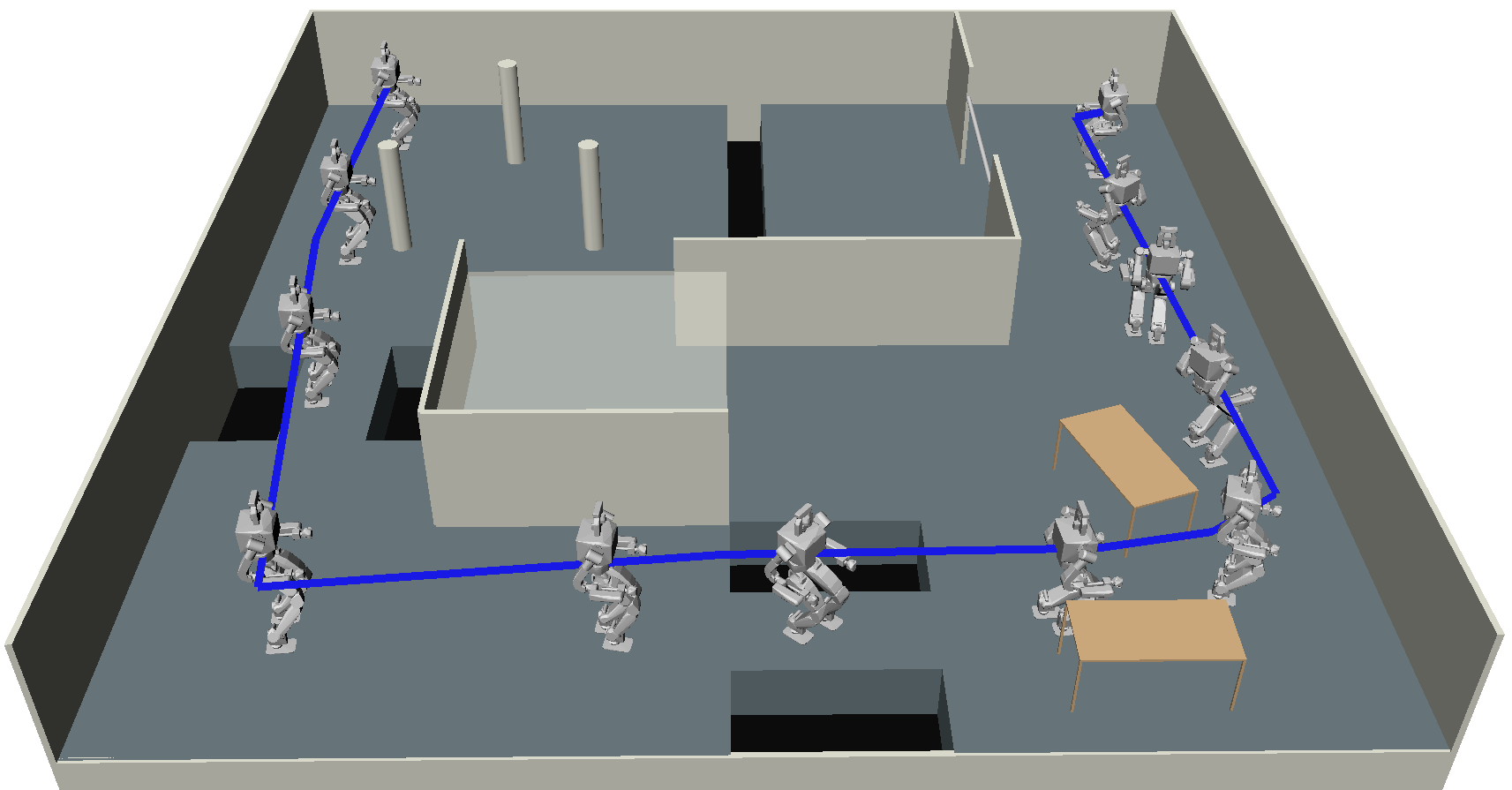}
  \hfil
  \includegraphics[width=0.48\textwidth]{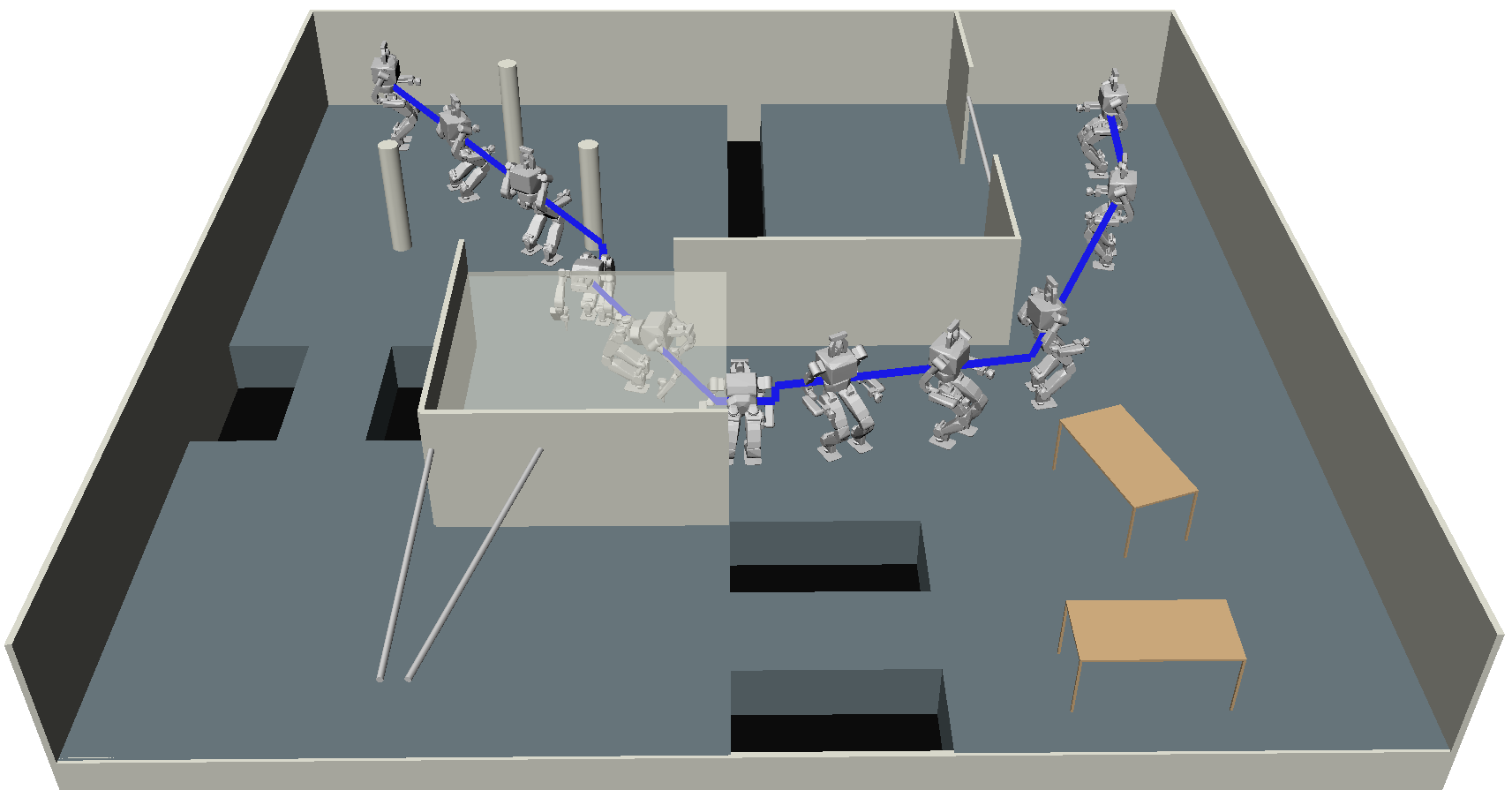}
  
  (a) \hfil \hfil (b)
  
  \vspace{2mm}
  
  \includegraphics[width=0.48\textwidth]{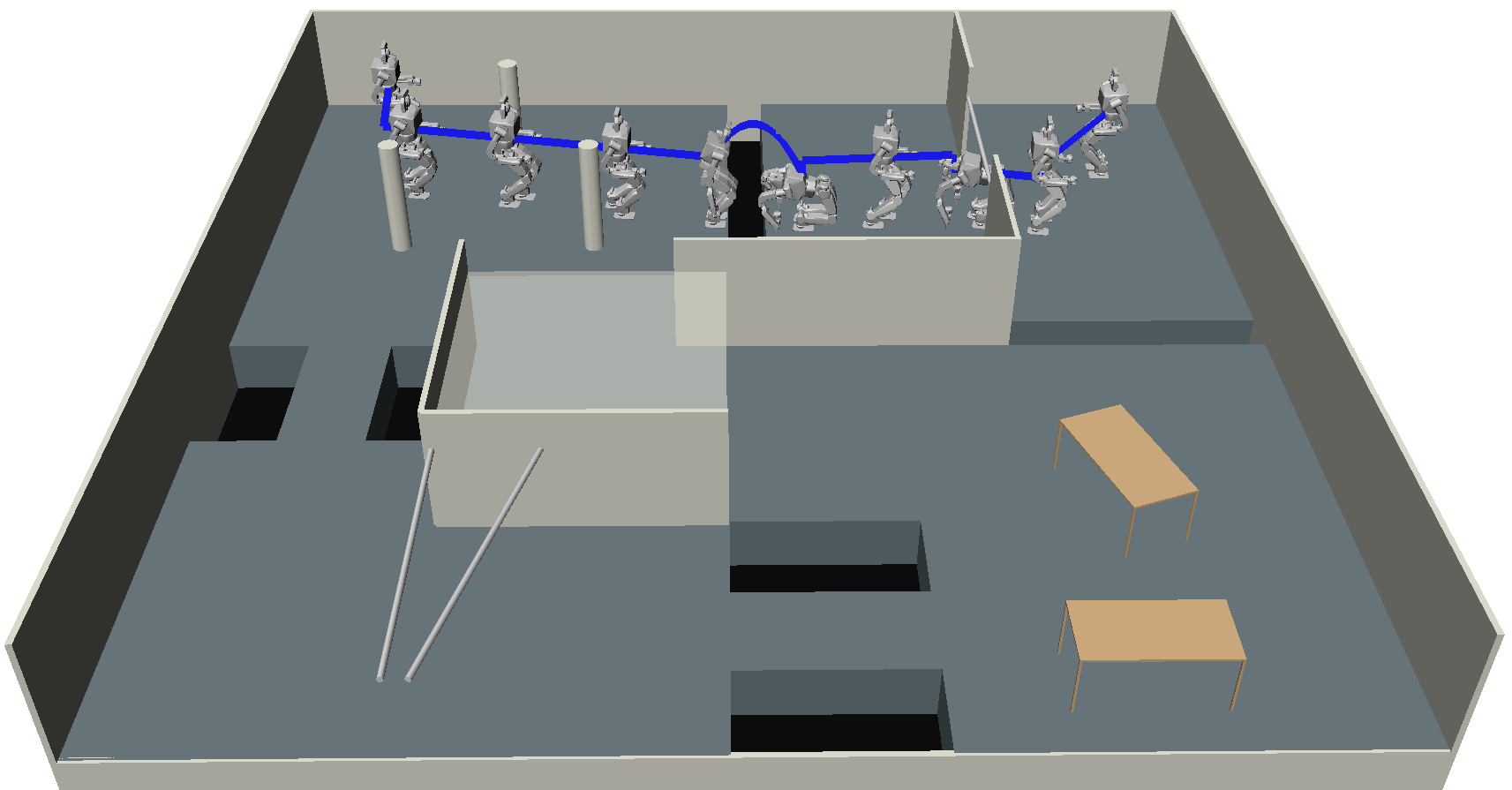}
  \hfil
  \includegraphics[width=0.48\textwidth]{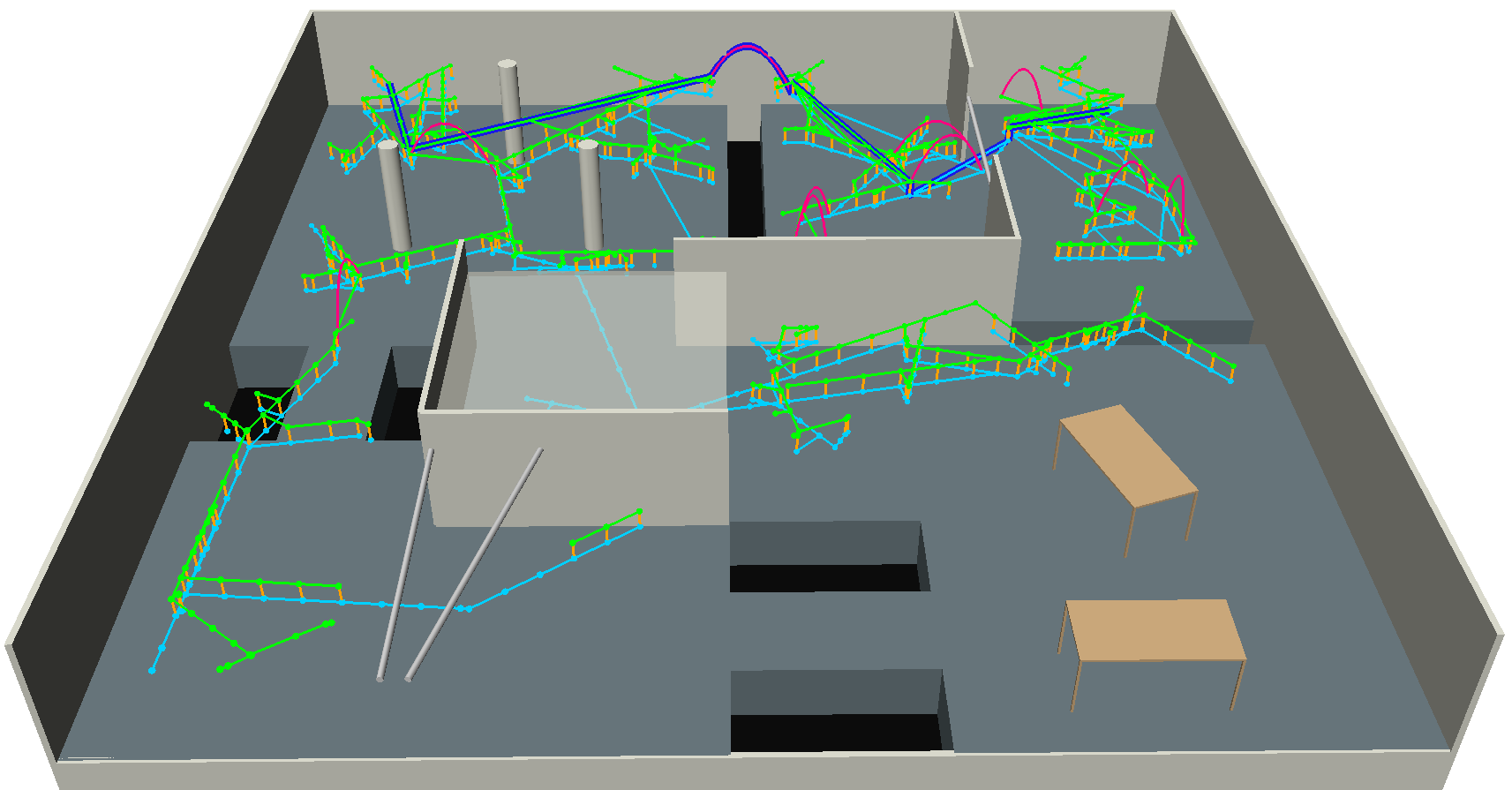}
  
  (c) \hfil \hfil (d)
  
  
  
  
  \caption{\label{fig:threeroutes} The three versions of the ``Three routes'' scenario. The robot must get from the back left corner to the back right corner. (a) A route exists that allows the robot to walk all the way to the goal. (b) Some bars were added to the walking route, so the robot must crawl at least once to reach the goal. (c) A gap was added at the end of the crawling routes, so the robot must jump at least once to reach the goal. (d) A grid that shows the map being explored.}
\end{figure}
\begin{figure}
  \centering
  \includegraphics[width=0.48\textwidth]{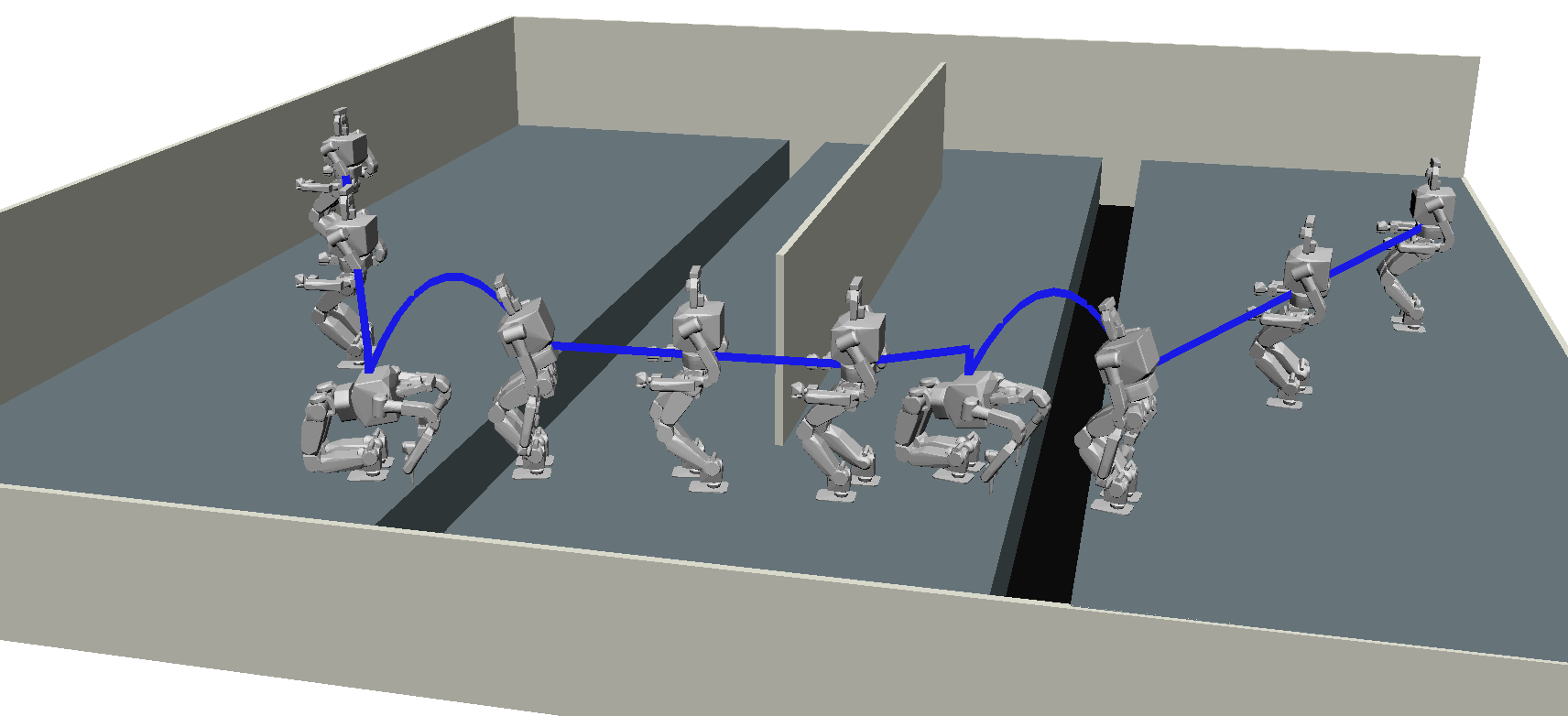}
  \hfil
  \includegraphics[width=0.48\textwidth]{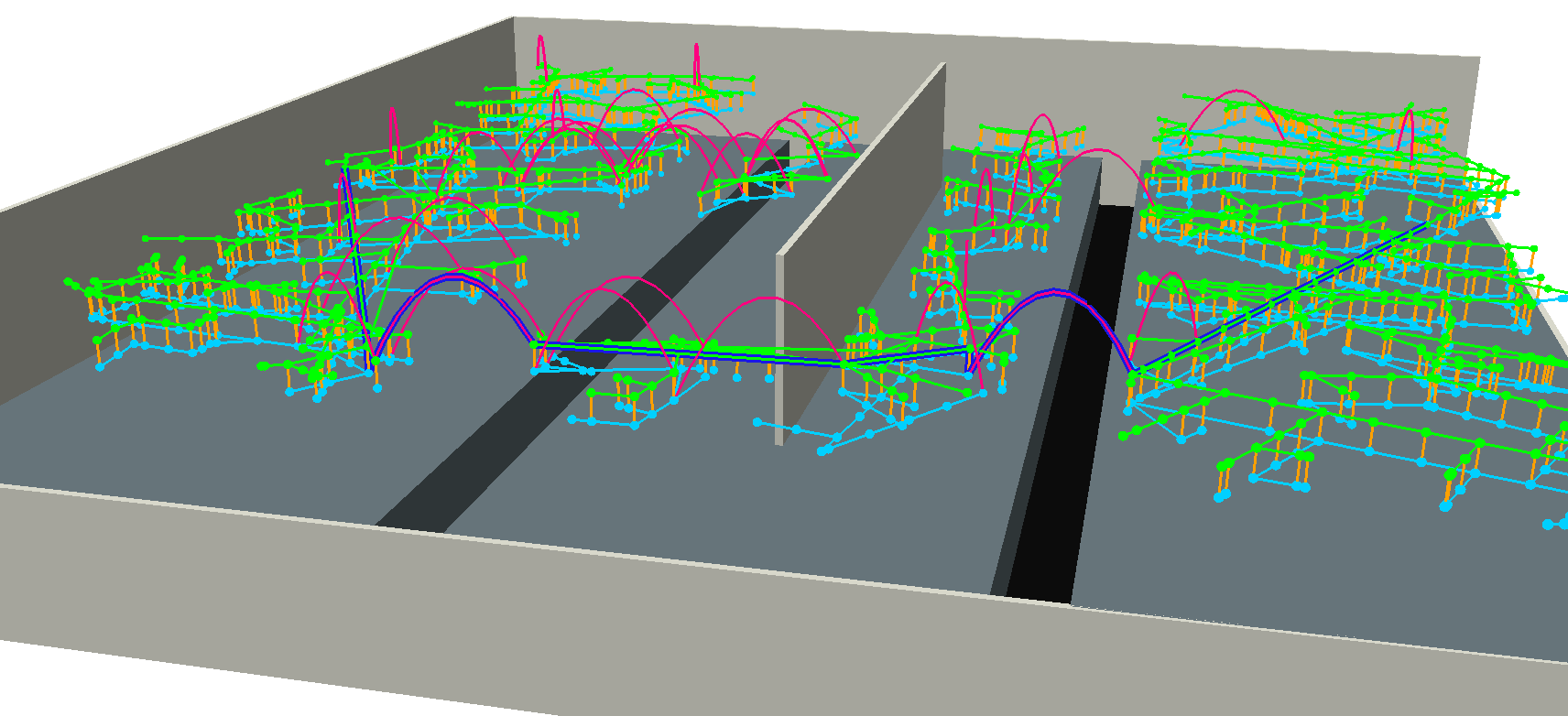}
  
  \caption{\label{fig:doublejump} The ``double jump'' scenario. The robot must jump across two gaps and navigate around a wall in the middle to get from the right side to the left side.}
\end{figure}

\begin{itemize}
  \item[] \textbf{Three Routes} scenario is shown in Fig. \ref{fig:threeroutes}. There are three potential routes that the robot might take to get from the start to the goal. We have three different versions of this scenario, and each version has progressively stronger requirements for what actions are needed by the solution, allowing us to compare the performance impact caused by specific action sequences being required.
  \item[] \textbf{Hallway} scenario was shown in Fig. \ref{fig:hallway}. The robot must crawl underneath some bars and then jump across a gap to get from the start on the right side to the goal on the left.
  \item[] \textbf{Double Jump} scenario is shown in Fig. \ref{fig:doublejump}. The robot must jump twice to get from the right side to the left.  
\end{itemize}

In Table \ref{tab:results} we see that the time required to solve a problem scales up with the number of actions being used (comparing the values in the \textbf{Graph} column of rows 1--3 and 4--5). For every action that is utilized by the planner, more exploration needs to be performed, which tends to increase the runtime. Not only does the action's space get explored, but also the transitions between the actions need to be explored. However, this cost is additive, not multiplicative, so the overall costliness will be related to the sum (not product) of the costliness of the individual actions. Jumping exploration is considerably more expensive than walking or crawling exploration. To curb this, we can modify Alg. \ref{alg:explore}, line \ref{alg:explore:growTowards} so that there is some probability of skipping the jumping expansion each cycle. In the results of Table \ref{tab:results}, we use a 90\% chance of skipping.

We can also see that the time required to solve a problem scales up with the number of actions \emph{required} by the environment to get a solution (comparing the \textbf{Graph} values of row 2 to 4 and of row 3 to 5). This is not surprising since requiring certain actions can be viewed as tightening the constraints on the solution, and tighter constraints tend to take longer to solve with randomized search.

\begin{table}
\caption{\label{tab:results}Time performance results, tested on an Intel\textsuperscript{\textregistered} Xeon\textsuperscript{\textregistered} Processor E3-1290 v2 (8M Cache, 3.70 GHz) with 16GB of RAM. $N_a$ is the number of action types that were provided to the planner. ``Graph'' is the time it took to generate a solved graph. ``Motion'' is the time it took to generate the physical motions for the solution. $\gamma$ is the ``Real-Time Ratio'', i.e. the time it would take to execute the plan divided by how long the whole plan (graph+motion) took to generate. ``Success Rate'' is how many times the planner succeeded (instead of timing out). All times are given in seconds. Each result is the average of 50 runs; the standard deviation is given in parentheses.}
\begin{tabular}{|l|c||l|l|l|r|}
\hline
\textbf{Scenario} & \textbf{$N_a$} & \textbf{Graph} & \textbf{Motion} & $\mathbf{\gamma}$ & \textbf{Success}\\\hline
Three Routes (a) & 1 & 0.088 (0.048) & 8.47 (0.81) & 143.2 (3.5) & 100\%\\\hline
Three Routes (a) & 2 & 0.134 (0.076) & 8.75 (0.91) & 143.6 (2.6) & 100\%\\\hline
Three Routes (a) & 3 & 0.484 (0.450) & 7.52 (1.86) & 136.8 (10.0) & 100\%\\\hline
Three Routes (b) & 2 & 0.152 (0.112) & 9.23 (1.09) & 142.5 (3.2) & 100\%\\\hline
Three Routes (b) & 3 & 0.561 (0.502) & 7.59 (2.30) & 134.0 (11.1) & 100\%\\\hline
Three Routes (c) & 3 & 1.210 (0.218) & 5.73 (1.79) & 121.2 (7.03) & 100\%\\\hline
Hallway          & 3 & 3.67 (11.52)  & 8.29 (0.84) & 127.2 (6.96) & 96\%\\\hline
Double Jump      & 3 & 1.48 (0.34)   & 4.32 (0.28) & 113.3 (6.24) & 100\%\\\hline
\end{tabular}
\end{table}

\subsubsection{Limitations}
Our experiments only used sufficient conditions for walking and crawling; any actions which violate the sufficient conditions for walking and crawling are ignored. Unfortunately, this eliminates the probabilistic completeness of the implementation. A more complete approach would consider the necessary conditions of walking and crawling, and then employ Multi-modal PRM \cite{hauser:ijrr2009} to examine walking/crawling segments where only the necessary conditions are met. This would allow the robot to step over small obstructions and squeeze through narrow passages between obstacles. These considerations will be the topic of future work.

\section{Conclusion}

We presented performance results of multi-action traversal plans being generated for the DRC-HUBO1 platform in complex environments. The complexity of the environments is derived from the fact that they require a variety of different action types to be interlaced in the correct sequence in order to navigate from the start to the goal. Three action types were used to traverse these environments: walking, crawling, and the standing long jump. The time required to fully generate the motion plans was less than 1/100th of the time that the motions require for physical execution. This makes the Possibility Graph a promising option for online use. Moreover, the time required to guarantee that a solution exists is even smaller, which suggests that the Possibility Graph would be an effective tool for higher-level task planners such as the Hybrid Backward-Forward planner \cite{garrett:hbf, grey:garrett:manip} which only needs to know whether a query is solvable.

The theoretical framework of the Possibility Graph can extend beyond the applications seen here. Future work will incorporate Multi-modal PRM to achieve probabilistic completeness in the quasi-static domain. We will also incorporate highly dynamic actions, e.g. running jumps, using nonlinear constrained optimization. This will open the door to fast, global, dynamic planning for high dimensional systems.


\begin{thebibliography}{1}

\bibitem{hsu:expansive}
Hsu, D., Latombe, J.-C.:
Path planning in expansive configuration spaces.
IEEE Int. Conf. on Rob. and Aut. (ICRA), vol. 3, 2719--2726 (1997)

\bibitem{berenson:tsr}
Berenson, D., Srinivasa, S. S., Kuffner, J.:
Task space regions: A framework for pose-constrained manipulation planning.
The Int. J. of Rob. Res. (2011)

\bibitem{garimort:footsteps}
Garimort, J., Hornung, A., Bennewitz, M.:
Humanoid navigation with dynamic footstep plans.
IEEE Int. Conf. on Rob. and Aut. (ICRA) 3982--3987 (2011)

\bibitem{hornung:anytime}
Hornung, A., Dornbush, A., Likhachev, M., Bennewitz, M.:
Anytime search-based footstep planning with suboptimality bounds.
12th IEEE-RAS Int. Conf. on Humanoid Rob. (Humanoids 2012) 674--679 (2012)

\bibitem{candido:hierarchical}
Candido, S., Kim, Y. T., Hutchinson, S.:
An improved hierarchical motion planner for humanoid robots.
8th IEEE-RAS Int. Conf. on Humanoid Rob. 654--661 (2008)

\bibitem{pettre:2stages}
Pettr\'{e}, J., Laumond, J. P., Siméon, T.:
A 2-stages locomotion planner for digital actors.
Proceedings of the 2003 ACM SIGGRAPH/Eurographics symposium on Computer animation 258--264 (2003)

\bibitem{dubois:possibility}
Dubois, D., Prade, H.:
Possibility Theory.
Meyers, R. A. (ed) Computational Complexity: Theory, Techniques, and Applications 2240--2252 (2012)

\bibitem{garrett:hbf}
Garrett, C. R., Lozano-P\'{e}rez, T., Kaelbling, L. P.:
Backward-forward search for manipulation planning.
IEEE/RSJ Int. Conf. on Intl. Rob. and Sys. 6366--6373 (2015)

\bibitem{grey:garrett:manip}
Grey, M. X., Garrett, C. R., Liu, C. K., Ames, A., Thomaz, A. L.:
Humanoid Manipulation Planning using Backward-Forward Search.
IEEE/RSJ Int. Conf. on Intl. Rob. and Sys. (2016) (to appear)

\bibitem{hauser:ijrr2009}
Hauser, K., Latombe, J. C.:
Multi-modal motion planning in non-expansive spaces.
The Int. J. of Rob. Res. (2009)

\bibitem{sentis:wholebody}
Sentis, L., Khatib, O.:
A whole-body control framework for humanoids operating in human environments.
Proceedings IEEE Int. Conf. on Rob. and Aut. 2641--2648 (2006)

\bibitem{sugihara:wholebody}
Sugihara, T., Nakamura, Y.:
Whole-body cooperative balancing of humanoid robot using COG Jacobian.
IEEE/RSJ Int. Conf. on Intl. Rob. and Sys., vol. 3, 2575--2580 (2002)

\bibitem{gienger:wholebody}
Gienger, M., Janssen, H., Goerick, C.:
Task-oriented whole body motion for humanoid robots.
IEEE-RAS Int. Conf. on Humanoid Rob. 238--244 (2005)

\bibitem{kuffner:connect}
Kuffner, J. J., LaValle, S. M.:
RRT-connect: An efficient approach to single-query path planning.
IEEE Int. Conf. on Rob. and Aut., vol. 2, 995--1001 (2000)

\bibitem{kavraki:prm}
Kavraki, L. E., Kolountzakis, M. N., Latombe, J.-C.:
Analysis of Probabilistic Roadmaps for Path Planning.
IEEE Trans. on Rob. and Aut., vol. 14, no. 1, 166--171, (1998)

\bibitem{zucker:chomp}
Zucker, M., Ratliff, N., Dragan, A.D., Pivtoraiko, M., Klingensmith, M., Dellin, C.M., Bagnell, J.A., Srinivasa, S.S.:
CHOMP: Covariant Hamiltonian optimization for motion planning.
The Int. J. of Rob. Res., 32(9-10), 1164--1193 (2013)

\bibitem{kuindersma:atlas}
Kuindersma, S., Deits, R., Fallon, M., Valenzuela, A., Dai, H., Permenter, F., Koolen, T., Marion, R., Tedrake, R.:
Optimization-based locomotion planning, estimation, and control design for the Atlas humanoid robot.
Autonomous Robots, 40(3), 429--455 (2016)

\bibitem{ayonga:hzd}
Hereid, A., Cousineau, E.A., Hubicki, C.M., Ames, A.D.:
3D Dynamic Walking with Underactuated Humanoid Robots: A Direct Collocation Framework for Optimizing Hybrid Zero Dynamics. 
IEEE Trans. on Rob. and Aut. (2016)

\end{thebibliography}
\end{document}